\documentclass[a4paper,fleqn]{cas-dc}
\usepackage[authoryear,longnamesfirst]{natbib}

\def\tsc#1{\csdef{#1}{\textsc{\lowercase{#1}}\xspace}}
\tsc{WGM}
\tsc{QE}

\usepackage{graphicx}

\usepackage{verbatim}
\usepackage{amssymb,amsmath,amsfonts,amsthm}
\usepackage{multirow}
\usepackage[linesnumbered,lined,boxed,commentsnumbered, ruled,vlined]{algorithm2e}
\usepackage{dsfont}
\usepackage{booktabs}
\usepackage{color, soul}
\usepackage{epsfig}
\usepackage{fancyhdr}
\usepackage{latexsym}
\usepackage{fancyvrb,relsize}
\usepackage{makeidx}
\usepackage{pstricks}
\usepackage{pst-node}
\usepackage{rotate}
\usepackage{rotating}
\usepackage{subcaption}

\usepackage{pstricks-add}
\usepackage{array}
\usepackage{url}

\usepackage{wrapfig}
\usepackage{makeidx}
\usepackage{tabularx} 
\usepackage{bm}
\usepackage{enumerate}
\usepackage{footnote}
\usepackage{relsize}
\usepackage{makecell}

\usepackage{booktabs, makecell}
\usepackage{siunitx}
\usepackage{longtable}
\usepackage{natbib}

\DeclareRobustCommand{\mythead}[1]{%
  \begin{tabular}{@{}l@{}}#1\end{tabular}%
}

\def\BibTeX{{\rm B\kern-.05em{\sc i\kern-.025em b}\kern-.08em
    T\kern-.1667em\lower.7ex\hbox{E}\kern-.125emX}}

\begin{document}
\let\WriteBookmarks\relax
\def\floatpagepagefraction{1}
\def\textpagefraction{.001}

\shorttitle{Multiplicative update rules for accelerating deep learning training and increasing robustness}    

\shortauthors{M. Kirtas, N. Passalis and A. Tefas$^1$}  

\title[mode=title]{Multiplicative update rules for accelerating deep learning training and increasing robustness}

\author[1]{Manos Kirtas}[orcid=0000-0002-8670-0248] \cormark[1] \ead{eakirtas@csd.auth.gr}
\author[1]{Nikolaos Passalis}[orcid=0000-0003-1177-9139] \ead{passalis@csd.auth.gr}
\author[1]{Anastasios Tefas}[orcid=0000-0003-1288-3667] \ead{tefas@csd.auth.gr}

\affiliation[1]{organization={Computational Intelligence and Deep Learning Group, Dept. of Informatics, Aristotle University of Thessaloniki}, city=Thessaloniki, country=Greece}

\begin{abstract}
Even nowadays, where Deep Learning (DL) has achieved state-of-the-art performance in a wide range of research domains, accelerating training and building robust DL models remains a challenging task. To this end, generations of researchers have pursued to develop robust methods for training DL architectures that can be less sensitive to weight distributions, model architectures and loss landscapes. However, such methods are limited to adaptive learning rate optimizers, initialization schemes, and clipping gradients without investigating the fundamental rule of parameters update. Although multiplicative updates have contributed significantly to the early development of machine learning and hold strong theoretical claims, to best of our knowledge, this is the first work that investigate them in context of DL training acceleration and robustness. In this work, we propose an optimization framework that fits to a wide range of optimization algorithms and enables one to apply alternative update rules. To this end, we propose a novel multiplicative update rule and we extend their capabilities by combining it with a traditional additive update term, under a novel hybrid update method. We claim that the proposed framework accelerates training, while leading to more robust models in contrast to traditionally used additive update rule and we experimentally demonstrate their effectiveness in a wide range of task and optimization methods. Such tasks ranging from convex and non-convex optimization to difficult image classification benchmarks applying a wide range of traditionally used optimization methods and Deep Neural Network (DNN) architectures.    
\end{abstract}


\maketitle

\section{Introduction}
\label{sec:introduction}

Even recent DL architectures are highly sensitive to training hyperparameters, initial weight, and data distributions~\cite{chaudhari2017entropysgd, 2102.06171}. This leads practitioners to convey a time-consuming search when their practical application is not fitted to the theoretical hypothesis of a designed architecture or optimization method. Over the last few years, several enhancements have been proposed to this end, ranging from variance-preserving~\cite{He_2015_ICCV} and data-driven initialization methods~\cite{9815870} to advanced optimization algorithms~\cite{adam} and clipping gradient techniques~\cite{1905.11881}.

However, building fast and stable optimization methods is still a challenging task that has been pursuing generations of researchers. Stochastic Gradient Descent (SGD) undoubtedly holds the credential of having a tremendous impact on the early development of DL and, despite its simplicity, remains a valuable solution in many challenging tasks~\cite{zhou2020towards}. Recently, \textit{adaptive learning rate} optimizers have been considered a choice in many applications, providing a great acceleration during training. Adagrad~\cite{adagrad} was the first optimizer to significantly outperform SGD when gradients are sparse or, in general, small. In turn, several efforts have been made to further improve optimization performance, with the variations of Adam~\cite{adam} and RMSProp~\cite{rmsprop} overcoming the performance degradation occurring in Adagrad when the loss function is non-convex and gradients are dense due to the rapid decay of learning rate~\cite{adam_beyond}. However, both Adam and RMSProp suffer from convergence issues related to the high variance in learning rate during the early stages of training~\cite{adptlr_beyond} or the magnitude of gradients~\cite{large_batch}, which provides, in turn, a fruitful area for analysis, variations, and enhancements~\cite{sun2019survey}.

In this work, we focus on the fundamental update rule, enabling us to investigate alternatives to optimization methods in a unified manner. Although multiplicative updates and optimization algorithms have been extensively studied during the early years of machine learning research~\cite{v008a006}, they have merely studied in context of DL learning optimization, making this work the first to demonstrate the practical benefits of applying them in DL optimization, with existing works in DL optimization focusing primarily on adapting the learning rate. Furthermore, there are strong theoretical claims, for example, by comparing the mistake bounds of the Winnow and Perceptron algorithms~\cite{KIVINEN1997325}, which show that multiplicative update rules leverage advantages over additive rules when a large portion of features are irrelevant~\cite{littlestone1988learning}, making it an excellent choice for the initial stage of training. Finally, Jeremy Bernstein et. al. present in~\cite{NEURIPS2020_9a32ef65} some preliminary results on employing multiplicative Adam in training lower precision neural networks representing model weights in a logarithmic number system, leading, however, to accuracy degradation in contrast to traditionally used optimizers when 32-bit resolution is applied, demonstrating mostly the potentiality of multiplicative updates. In contrast to~\cite{NEURIPS2020_9a32ef65}, in this work we propose a Generic Optimization Framework for Alternative Updates (GOFAU) in order to investigate multiplicative updates in a wide range of optimization algorithms, while we demonstrate that with the appropriate proposed update rules, neural networks training can be significantly accelerated and leads to a more robust training procedure.     

To this end, we propose a novel framework for alternative update rules that can be used out-of-the-box, incorporating the already known techniques, such as adaptive learning rate. More specifically, the proposed framework leverages advantages over the traditional used update rules in terms of robustness and acceleration in the first stages of training. It is motivated by the fact that the multiplicative update rule takes into account the magnitude of parameters increasing robustness and accelerating the convergence rate since it scales parameters according to their magnitudes. Additionally, it normalizes gradients while integrating the existing benefits of traditionally used optimizers. More precisely, the proposed update function provides a simple measurement of how much a single gradient descent step will scale the original parameter and naturally normalizes gradients, ignoring the gradient's magnitude and considering the parameter's one, which empirically has shown that makes training more robust and faster~\cite{large_batch, Bachlechner2021}.    

Note that training with multiplicative updates constraints parameters to their initial sign. However, this sign limitation has insignificant negative effects on neural networks performance, especially in the first training epochs as experimentally demonstrated, in which the parameters are normally distributed and they are over-parameterized by default. To overcome this sign limitation, we extend the proposed multiplicative update method to a hybrid multiplicative-additive update rule that allows parameters to change sign while exploiting the robustness and acceleration capabilities of the multiplicative update term. However, it is worth mentioning that the proposed multiplicative update can be especially useful in training monotone~\cite{5443743} and nonnegative neural networks oriented to explainability~\cite{8051252, LEMME2012194} as well as in neuromorphic neural networks~\cite{moralis2022neuromorphic, pleros2021compute}.    

The main contributions of this paper are a) a generic framework for alternative update rules that works out-of-the-box for all well-known optimization methods, b) a multiplicative update rule that accelerates convergence of the optimization process while making models more robust to initial parameter distribution, and c) a hybrid multiplicative-additive update rule that overcomes the sign limitation of the multiplicative update term. We first demonstrate the capabilities of the proposed framework in 2-dimensional spaces with both convex and non-convex optimization tasks, highlighting its advantages in easily visualized and understandable spaces. In turn, we report experimental results on traditional image classification benchmarks, such as CIFAR and Tiny ImageNet, employing the proposed framework in several optimization algorithms.

The rest of the paper is structured as follows. First, we introduce the proposed framework in Section~\ref{sec:proposal}. Then, the results of the experimental evaluation are reported in Section~\ref{sec:experiments}. Finally, the conclusions are drawn in Section~\ref{sec:conclusions}.
\section{Proposed method}
\label{sec:proposal}

First, we present a proposed generic optimization framework that enables one to apply alternative update rules. In turn, using the proposed framework, we introduce a novel multiplicative update rule and a hybrid additive-multiplicative optimization method that leverages advantages over traditional optimization methods in terms of acceleration and robustness.

\subsection{Generic Optimization Framework for Alternative Updates}

We propose a Generic Optimization Framework for Alternative Updates (GOFAU), inspired by~\cite{adam_beyond}, to present multiplicative iterative optimization methods in an online optimization problem with a full information feedback setting. We use the online optimization setup due to its flexibility and simplicity, providing us with an easy way to present alternative update rules applied in a wide range of optimization algorithms. It should be noted that without loss of generality, the proposed framework can be easily applied to mini-batch optimization as well. 

\begin{algorithm}[htb]
  \SetKwData{Left}{left}\SetKwData{This}{this}
  \SetKwData{Up}{up}\SetKwFunction{Union}{Union}
  \SetKwFunction{FindCompress}{FindCompress}
  \SetKwInOut{Input}{Input}
  \SetKwInOut{Output}{Output}
  
  \Input{\(\eta\): step size, \\
    \(\phi(\cdot)\): calculates momentum,\\
    \(\psi(\cdot)\) calculates adaptive learning rate,\\
    \(\xi(\cdot)\) update rule,
    \(\theta_0\): initial parameter,\\
    \(f(\theta)\): stochastic objection function
  }
  \Begin{
      \While{\(t=1\) to \(T\) }{
        \(g_t = \nabla_{\theta}f_t(\theta_{t-1})\)  \;
        \(m_t = \phi_{t}(g_1,\ldots, g_t) \) \;
        \(l_t = \psi_{t}(g_1,\ldots, g_t) \) \;
        \(\Delta\theta_t = \xi(\theta_{t-1}, m_t, l_t, \eta)\)
      }
    }
    \caption{Generic Optimization Framework for Alternative Updates (GOFAU) \label{alg:optim}}    
\end{algorithm}

According to this online setup, at each time step \(t\), the gradient-based algorithm picks a point (i.e. the parameter \(\theta\) of the model to be learned) \(\theta_t \subset \mathcal{F}\), where \(\mathcal{F} \in \mathbb{R}^{n}\) is the feasible set of points. The loss function \(f_t\) for time step \(t\) is revealed (to be interpreted as the loss of the model with the parameter chosen in the next mini-batch), with the algorithm loss being \(f_t(\theta_t)\). In the simplest case of online SGD, the algorithm moves the point \(\theta_t\) in the opposite direction of the gradient \(g_t=\nabla f_t(\theta_t)\). To adequately describe adaptive learning rate algorithms as well, we use Algorithm~\ref{alg:optim}, which is a generic framework that can describe various popular stochastic gradient descent algorithms allowing one to apply alternative updates. More specifically, the optimization procedure in the given framework can be specified by different choice of \(\phi(\cdot)\) and \(\psi(\cdot)\), where specifies how the momentum, \(m_t\), and the adaptive learning rate, \(l_t\), are calculated at the time step \(t\), respectively. Furthermore, the \(\xi(\cdot)\) choice defines the update rule of the optimization algorithm. Thus, given the generic optimization framework, the SGD algorithm can be described as:
\begin{equation}
  \label{eq:sgd}
  \begin{gathered}
    \phi_{t}(g_1,\ldots, g_t) = g_t, \\ 
        \psi_{t}(g_1,\ldots, g_t) = \mathbf I_t.
    \end{gathered}
\end{equation}

The adaptive optimization methods aim to preserve convergence by choosing the appropriate averaging functions. Adagrad adapts the learning rate for each feature depending on the estimated geometry of the problem and it uses the following functions:
\begin{equation}
\label{eq:adagrad}
\begin{gathered} 
    \phi_{t}(g_1,\ldots, g_t) = g_t, \\
    \psi_{t}(g_1,\ldots, g_t) = \sqrt{\frac{1}{{diag\left(\sum_{i=1}^{t}g_i^2\right)}}},
\end{gathered}
\end{equation}
where \(diag(\cdot)\) denotes the diagonal elements of a matrix.

Modern variants replaced the simple averaging function of the adaptive learning rate function, \(\psi(\cdot)\), which Adagrad uses, with an exponential moving average (EMA) function. Adam and RMSprop fall into this category. Adam, for example, calculates the EMA of the gradient and square gradient with hyperparameters \(\beta_1, \beta_2 \in [0, 1)\) used to control the decay rates of these moving averages. Given the generic optimization algorithm, Adam is formulated as: 
\begin{equation}
\label{eq:adam}
\begin{gathered} 
    \phi_{t}(g_1,\ldots, g_t) = \frac{(1-\beta_1)\sum_{i=1}^t\beta_1^{t-1}g_i}{1-\beta_1^t}, \\
    \psi_{t}(g_1,\ldots, g_t) = \sqrt{\frac{1-\beta_2^t}{(1-\beta_2)\sum_{i=1}^{t}\beta_2^{t-1}g_i^2}}.
\end{gathered}
\end{equation}
This update can alternatively be stated by the following simple recursion:
\begin{equation}
    \label{eq:adam_alt}
    \begin{gathered}
    m_{t,i} =\beta_1 m_{t-1,i} + (1-\beta_1)g_{t,i} \\
    l_{t,i} =1 / \left( \sqrt{\beta_2 l_{t-1,i} + (1-\beta_2)g_{t,i}^{2}} + \epsilon\right)
    \end{gathered}
\end{equation}
where \(m_{0, t}=0\) and \(l_{0, i}=0\) for all \(i\in[d]\), and \(t \in [T]\). The \(\epsilon=10^{-8}\) denotes a small constant value ensuring numerical stability, while the values of \(\beta_1=0.9\) and \(\beta_2=0.99\) are typically recommended in practice.  Finally, RMSProp can be expressed as the variance of Adam by setting \(\beta_1=0\).

The common of all the aforementioned and traditionally used optimization methods is that they apply additive update rule, meaning that they add the update vector to the parameter's vector without taking into account the magnitude of the parameter. The general update rule given the momentum and adaptive learning rate is given by:
\begin{equation}
    \label{eq:normal_update}
    \xi(m_t, l_t) = \eta m_t l_t.
\end{equation}

\subsection{Multiplicative Updates}

The additive update rule depends on the gradient values and the learning rate, making it extremely sensitive to the magnitude of the parameters, especially when it is far from an optimal local minima. This profoundly exists when the initialization is bad or in cases where the learning rate is set to the default value, but the magnitude of parameters differs from the initial theoretical hypothesis. To overcome these limitations, we propose a multiplicative update rule oriented for faster convergence and robustness that normalizes and clips gradients. More specifically, the proposed update rule incorporates \(\tanh(\cdot):\mathbb{R}\rightarrow(-1,1)\) function that offers normalization of the gradient and then multiplies it by the parameter. In this way, the proposed update term proportionally scales the parameters considering the magnitude of it. Exploiting the proposed GOFAU, we proposed a multiplicative update rule given by:
\begin{equation}
    \label{eq:m_abs}
    \xi(m_t, l_t) = |\theta_{t-1}|\tanh{\left(\eta_{in} m_t l_t \right)}\eta_{out},
\end{equation}
where \(\eta_{in} \in \mathbb{R}^{+}\) is the inner and \(\eta_{out} \in (0,1]\) the outer learning rate. Essentially, the inner learning rate allows one to adjust the gradients regarding the working range of the used nonlinearity. The outer learning rate affects the size of the step similarly to the learning rate used in traditionally applied optimization methods and, additionally, defines the upper scaling threshold depending on the parameter, \(\left[-\eta_{out}|\theta_{t-1}|, \eta_{out}|\theta_{t-1}|\right]\).

Motivated by the observation that gradient clipping in a specific setting can accelerate training~\cite{1905.11881}, making DL models more robust~\cite{large_batch}, the proposed multiplicative update rule naturally normalizes gradients and introduces a threshold proportional to the magnitude of the parameter. More specifically, the proposed update rule makes the update term proportional to the parameter by introduction \(\tanh(\cdot):\mathbb{R}\rightarrow(-1,1)\) function.  Intuitively, the multiplier \(\tanh{\left(\eta_{in} m_t l_t \right)}\) provides a measurement of how much a single gradient descent step will scale the original parameter. In this way, the divergence issues occurred when update term \(||\eta\nabla J(\bm{w}_t)||\) becomes significantly larger than the weight \(||\bm{w}||\) can be partially eliminated, where \(||\cdot||\) denotes the L2 norm. 

Indeed, as already shown, the ratio of the L2-norm of weights and gradients, \(||\bm{w}||/||\nabla L(\bm{w}_t)||\), is not only significantly high in the first epochs of training, but also highly different between weights, biases and layers~\cite{1908.03265}. As a result, vanishing and exploiting gradient phenomena are prevalent during the initial training stage, making the traditional optimizer highly sensitive to initialization and learning rate~\cite{1708.03888}. Using the proposed method, the updated weight no longer depends on the magnitude of the gradient, preventing the gradient-weight ratio \(||\bm{w}||/||\nabla L(\bm{w}_t)||\) to become significantly large, while it introduces thresholds for maximum increment and decrement that can be easily controlled by the outer learning rate. Thus, we claim that the proposed update term makes training more robust to vanishing and exploiting gradience phenomena~\cite{1708.03888}. 

Furthermore, the proposed update rule has the ability to preserve the initial sign of the update parameter, making it an excellent choice for training neural networks oriented to interpretability~\cite{8051252, 6783731} and neuromorphic architectures~\cite{9605987}. Several works have shown that non-negativity constraints leverage advantages to the interpretability of models, since they allow one to perform part-based learning due to the elimination of canceling neurons, resulting in additive data representation~\cite{7310882, LEMME2012194} of models. Therefore, non-negative models provide further insights that can be used to improve the performance of DL models~\cite{8489216}. Last but not least, constraining parameters on the initial sign can be especially useful in the case of neuromorphic architectures, since it allows one to know beforehand the sign of each parameter. In this way, complex sign mechanisms can be replaced or simplified, leading to significantly lower hardware complexity~\cite{Totovic_2022}.

However, in traditional DL learning training, there are no such limitations. Still, the current neural network architectures are initialized by drawing parameters from a normal distribution, \(\mathcal{N}(0, \sigma^2)\), with zero mean and variance that depend on the size of the layers~\cite{He_2015_ICCV}. Additionally, DL architectures by default are overparameterized. This allows them to converge even in complex loss topologies, stabilize and accelerate training, and, in turn, improve overall performance~\cite{pmlr-v80-arora18a}. At the same time, overparameterization allows different optimization methods to converge in different local minima without significant performance degradation~\cite{1412.6544, 2104.05605}. In this work, we exploit overparameterization to apply multiplicative updates that accelerate training, since it allows the proposed multiplicative update rule to converge in a local minimum using the initial sign of the parameters. More specifically, such proportional to magnitude scaling of parameters, normalization and clipping of gradients that multiplicative update term facilitates allowing the different parameters of the overparameterized network to adaptively converge to an optimal faster, in contrast to additive term in which the updates are made only by considering gradient and as result making updates highly related to loss functions. This leads a great amount of parameters to perform additional corrections steps until the majority of parameters converge to their suboptimal values, slowing down the convergence when additive updates are used.

Where needed, the multiplicative update rule can be combined with the additive one, exploiting in this way the benefits of the multiplicative update rule with the ability of parameters to change sign when additive rule is used. This is especially useful in cases where the model's capacity is small.  The proposed hybrid rule retains the advantages of multiplicative updates while controlling the contribution of each update term. The proposed hybrid multiplicative additive update rule is given by:
 \begin{equation}
     \label{eq:hybrid}
     \xi(m_t, l_t) = \gamma\left(|\theta_{t-1}|\tanh{\left(\eta_{in} m_t l_t \right)}\eta_{out}\right) + (1-\gamma) \left(\eta m_t l_t\right), 
 \end{equation}
 where \(\gamma\) is the weight of the relative contribution of the multiplicative term. 

Finally, the hybrid method, except for allowing parameters to change their initial sign when using the multiplicative update rule, also ensures that the parameters will not be stuck at zero, which is a potential consequence of utilizing multiplicative update rules. Although modern DL frameworks, such as PyTorch~\cite{paszke2017automatic} and Tensorflow~\cite{tensorflow2015-whitepaper}, do not initialize weights and biases to zero, combining the multiplicative update term with the additive one, allows one to overcome such potential limitation, leveraging the advantages of the multiplicative update. However, it should be mentioned that eliminating synapses could potentially be useful in cases where weight sparsity and/or pruning are required~\cite{7310882}, while it can also provide an additional regularization effect, avoiding in this way the overfitting during training.             
\section{Experimental Evaluation}
\label{sec:experiments}

We experimentally evaluated the proposed framework using two different sets of experiments. First, we demonstrate the effectiveness of the proposed optimization framework in 2-dimensional convex and non-convex tasks, giving us further insights into the optimization process. Then, we apply the proposed optimization method in traditionally used image classification benchmarks employing DNNs. 

\subsection{Convex and non-convex optimization}

For a better understanding of the optimization method under the proposed framework, we demonstrate different optimizers in 2-dimensional convex and non-convex tasks, which can be easily visualized. For convex task, we used a second order polynomial given by:
\begin{equation}
    \label{eq:2d_convex}
    f_{1}(\bm{x}) = \beta (x_1-\alpha)^2 + 10\beta(x_2-\alpha)^2, 
\end{equation}
where \(\bm{x}^*=[\alpha,\alpha]\) is the global minimum of the convex function with subscript \(i\) denoting the \(i\)-th element of the vector, and \(\beta\) defines the steepness of the function. Additionally, we use the non-convex Rosenbrock benchmark function, given by:
\begin{equation}
    f_2(\bm{x}) = (\alpha - x_1) + \beta(x_2-x_1^2)^2,
\end{equation}
where the global minimum is at \(\bm{x}^* = [\alpha, \alpha]\) and \(\beta\) defines the steepness of the function. The \(\alpha\) and \(\beta\) values are set to 1 and 20, respectively. The vector parameter \(\bm{x}\) is optimized to solve the equations. While for the evaluation, we report the Euclidean distance between the parameters and the global minimum. 

\begin{table}[]
    \centering
    \caption{Task configurations for tuning and evaluation processes}
    \label{tab:tune_config}
    \begin{tabularx}{\linewidth}{X|XX|XX}
         \toprule
         &\multicolumn{2}{c|}{\textbf{Convex 2D}} & \multicolumn{2}{c}{\textbf{Rosenbrock}} \\
         \cmidrule{2-5}
         Parameter & Tuning & Evaluation & Tuning & Evaluation \\ \hline
         \(x_1^{(t=0)}\) & \(50\) & \(\mathcal{N}(50, 5)\) & 0.5 & \(\mathcal{N}(0.5, 0.1)\)\\
         \(x_2^{(t=0)}\) &  \(50\) & \(\mathcal{N}(50, 5)\) & 3.0 & \(\mathcal{N}(3, 1)\)\\
         \(\alpha\) & \(1\) & \(\mathcal{N}(1, 1)\)  & 1 & 1\\
         \(\beta\) & \(20\) & \(\mathcal{N}(20, 2)\) & \(60\) & \(\mathcal{N}(60, 6)\)\\ 
         Iterations & \(100\) &   \(\mathcal{N}(100, 10)\) & 100 & \(\mathcal{N}(100, 10)\) \\
         \hline
    \end{tabularx}

\end{table}

Firstly, we tune the hyperparameters of each evaluated update rule on the aforementioned tasks with a given configuration. The configuration used for each optimization task is reported in the Table~\ref{tab:tune_config} at columns 2 and 4. For hyperparameter tuning, a grid search method is used. More specifically, for the traditionally used additive update rule, we perform a search for the learning rate, ranging from \([1e-6, 5e+2]\), with a step of \(0.5\). For the proposed multiplicative update rule that facilitates the proposed GOFAU, we performed a grid search for inner and outer learning rates, between values \([1e-1, 5e+1]\) and \([1e-4, 1.0]\), using the step \(0.5\), respectively.  Finally, for the hybrid variation of the proposed framework, the above learning rate search spaces are applied, while the weight of the relative contribution of the multiplicative term, \(\gamma\), is kept fixed and equal to \(0.5\).

The performance of the best configurations obtained from the tuning process is presented in Table~\ref{tab:tune_convex}. More specifically, we report the Euclidean distance from global minimum after 100 iterations using different optimizers and update rules in different optimization tasks. In the second column of the table, we report the distance when we apply the traditionally used optimization method. In columns three and four, we report the performance when we apply the proposed GOFAU applying multiplicative and hybrid update rules, respectively. The proposed methods are denoted in tables using bold text. In rows 1 and 6, in addition to the names of tasks, we also report the Euclidean distance between the global minimum and the initial point. 

\begin{table}[]
    \centering
     \caption{Distance from global minimum after 100 iterations using the best training configuration}
    \label{tab:tune_convex}
    \begin{tabularx}{\linewidth}{l|l|l|l}
         \toprule
         Optimizer & \mythead{\normalfont Baseline\\\normalfont(Additive)} & \mythead{\textbf{GOFAU}\\\textbf{(Multiplicative)}} & \mythead{\textbf{GOFAU}\\\textbf{(Hybrid)}} \\
         \midrule
         \multicolumn{4}{c}{\textbf{Convex 2D} \textit{(69.26)}}\\ \midrule
         SGD & \(8.61\times10^{-1}\) & \(2.92\times10^{-5}\) & \(\bm{3.57\times10^{-7}}\)  \\
         Adagrad & \(5.80\times10^{-4}\) & \(1.35\times10^{-4}\) & \(\underline{\bm{1.68\times10^{-7}}}\) \\
         Adam & \(3.75\times10^{-2}\) & \(7.44\times10^{-3}\) & \(\bm{4.47\times10^{-3}}\)  \\
         RMSProp & \(1.82\times10^{-5}\) & \(7.58\times10^{-6}\) & \(\underline{\bm{1.68\times10^{-7}}}\)  \\ \midrule
         \multicolumn{4}{c}{\textbf{Rosenbrock} \textit{(2.06)}}\\ \midrule
         SGD & \(1.38 \times 10^{0}\) & \(7.70\times10^{-2}\) & \(\bm{2.04\times10^{-2}}\) \\
         Adagrad & \(4.60 \times 10^{-1}\) & \(1.88\times10^{-1}\) &  \(\underline{\bm{1.26\times10^{-2}}}\) \\
         Adam & \(8.24\times 10^{-1}\) & \(7.84\times10^{-2}\) & \(\bm{5.73\times10^{-2}}\) \\
         RMSProp & \(4.58\times10^{-1}\) & \(1.85\times10^{-1}\)  & \(\bm{5.06\times10^{-2}}\)  \\
        \bottomrule
    \end{tabularx}
  
\end{table}

The experimental results highlight that, with the appropriate hyperparameter tuning, the proposed multiplicative update rule consistently improves the performance of applied optimizers. We observe that the greater improvements are obtained in the non-adaptive learning rate optimization method, such in the SGD, and especially when the distance from the global minimum is large, such in the Convex 2D task. This is an expected behavior since the multiplicative update rule depends on the parameter's magnitude, and combined with the normalization that is offered, it converges faster to global minimum. At the same time, traditional SGD suffers from convergence issues, since the gradients in the first iterations are large, while when it reaches closer to the global minimum, the gradients become smaller, making the choice of an optimal learning rate difficult or even impossible. This issue is traditionally solved with adaptive learning rate methods that enable optimizers to use bigger steps when they are far from a local (or global) minimum and gradient magnitudes are larger. However, we observe that even in the adaptive learning rate methods, the optimizers still fall in worse minimums in contrast to the multiplicative updates. Furthermore, it is shown that optimal performance is obtained when using the proposed GOFAU by combining the multiplicative update rule with the additive one under the proposed hybrid method. In conclusion, the proposed hybrid update rule significantly outperforms both the additive and multiplicative update rules.

\begin{figure}
    \centering
    \includegraphics[width=\linewidth]{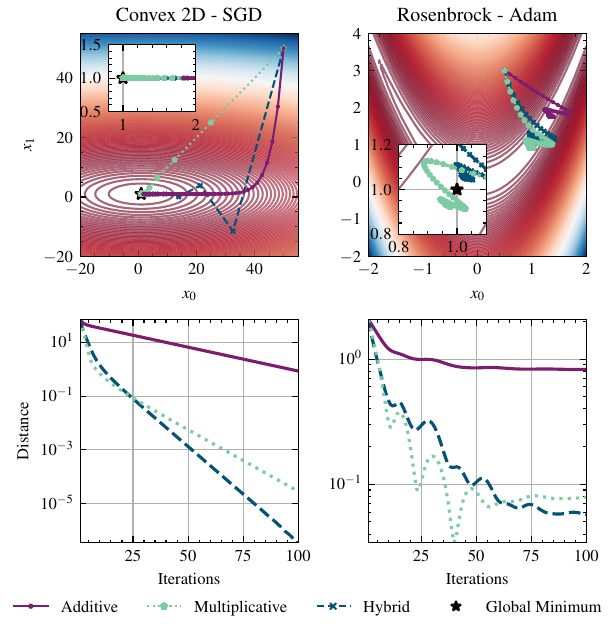}
    \caption{The top figures demonstrate the optimization process for Convex 2D and Rosenbrock tasks alternative updates for SGD and Adam optimizers, respectively. The bottom figure reports the Euclidean distance between the parameters and the actual global minimum.}
    \label{fig:convex_opt}
\end{figure}

\begin{table*}
    \centering
    \caption{Evaluation of optimizers on 100 randomly drawn configurations}
    \label{tab:convex_res}
    \begin{tabularx}{\textwidth}{l|X|X|X}
         \toprule
         Optimizer & \mythead{\normalfont Baseline\\\normalfont(Additive)} & \mythead{\textbf{GOFAU}\\\textbf{(Multiplicative)}} & \mythead{\textbf{GOFAU}\\\textbf{(Hybrid)}} \\
         \midrule
         \multicolumn{4}{c}{\textbf{Convex 2D}}\\ \midrule
         SGD & \(1.48\times10^{-2}\pm{8.06\times10^{-3}}\) & \(1.82\times10^{-3}\pm{5.17\times10^{-3}}\) & \(\bm{8.37\times10^{-4} \pm{2.35\times10^{-3}}}\)  \\
         Adagrad & \(7.61\times10^{-5}\pm{1.67\times10^{-4}}\) & \(2.30\times10^{-3}\pm{7.17\times10^{-3}}\) & \(\underline{\bm{1.79\times10^{-9} \pm{1.34\times10^{-9}}}}\) \\
         Adam & \(4.28\times10^{-3}\pm{4.28\times10^{-3}}\) & \(3.95\times10^{-3}\pm{4.88\times10^{-3}}\) & \(\bm{1.26\times10^{-3}\pm{1.45\times10^{-3}}}\)  \\
         RMSProp & \(8.26\times10^{-6}\pm{2.33\times10^{-5}}\) & \(2.24\times10^{-3}\pm{2.24\times10^{-3}}\) & \(\bm{2.01\times10^{-8}\pm{9.58\times10^{-8}}}\)  \\ \midrule
         \multicolumn{4}{c}{\textbf{Rosenbrock}}\\ \midrule
         SGD & \(5.88\times10^{-1}\pm{2.19\times10^{-1}}\) & \(1.75\times10^{-1}\pm{2.36\times10^{-1}}\) & \(\underline{\bm{7.42\times10^{-2}\pm{1.53\times10^{-1}}}}\) \\
         Adagrad & \(3.53\times10^{-1}\pm{3.59\times10^{-1}}\) & \(2.03\times10^{-1}\pm{2.13\times10^{-1}}\) &  \(\bm{1.98\times10^{-1}\pm{2.68\times10^{-1}}}\) \\
         Adam & \(4.12\times10^{-1}\pm{1.69\times10^{-1}}\) & \(1.68\times10^{-1}\pm{2.55\times10^{-1}}\) & \(\bm{1.48\times10^{-1}\pm{1.58\times10^{-1}}}\) \\
         RMSProp & \(3.16\times10^{-1}\pm{3.08\times10^{-1}}\) & \(1.43\times10^{-1}\pm{1.67\times10^{-1}}\)  & \(\bm{2.01\times10^{-1}\pm{2.51\times10^{-1}}}\)  \\
        \bottomrule
    \end{tabularx}
\end{table*}

To better understand the differences of each optimization process, we plot two cases in Figure~\ref{fig:convex_opt}. On the first column, the optimization process is depicted when applying the SGD optimizer on the Convex2D task. In this case, the multiplicative update rule significantly accelerates training during the first 20 iterations, outperforming both the additive and hybrid update rules. After 20 iterations, the multiplicative method moves points slower to global minima than the hybrid update rule. Similar behavior is also depicted in the second column of the figure, demonstrating the Rosenbrock optimization task that is optimized with the Adam method. We suspect that such a slower convergence in the last iterations of training comes from the fact that both the parameters and the gradients are small in magnitude. The combination of additive and multiplicative update rules overcomes this issue and significantly outperforms the additive baseline.  


To investigate the robustness of the proposed framework, we applied the best hyperparameter configurations obtained in different setups of optimization tasks. More precisely, we draw the initial points, \(\beta\) value (which defines the slope of the function), and the number of iterations of a Gaussian distribution, with a mean equal to the value used in the tuning process and a standard deviation that is proportional to the magnitude of the mean, as reported in Table~\ref{tab:tune_convex}. In addition, in the Convex2D case, the global minimum is drawn from a normal distribution as well. In Table~\ref{tab:convex_res}, we present the average and standard deviation of scores over 100 evaluation runs for the three update rules using randomly drawn task configurations. The score is computed by dividing the Euclidean distance after the last iteration by the initial distance. In this way, the zero means that the parameters are on the global minimum. 

In the Convex 2D case, the improvements when the hybrid update rule is used are impressive contrary to the traditionally used updates. More specifically, the hybrid update function significantly improves the performance for both non-adaptive and adaptive learning rate algorithms. Adagrad achieves the best performance in the average case, with the proposed hybrid update rule improving the performance of the traditional optimizer by 5 orders of magnitude. The performance of multiplicative updates is constantly near to \(10^{-3}\), highlighting in this way the limitations of multiplicative updates when a sign change is required. The performance of hybrid update rule demonstrates that not only sign limitation can be overcome, but also that by combining the multiplicative update with additive one can significantly improve the performance of models in terms of robustness to the initial parameters. 

Similar results are obtained also for the Rosenbrock non-convex optimization task, with hybrid update constantly achieving better performance, highlighting its generalization ability in non-convex tasks even in cases where models are not overparameterized. Furthermore, the robustness and generalization ability of the proposed framework is demonstrated, achieving better performance than the traditional updates in cases where the initial theoretical hypothesis differs from the actual one. In this task, hybrid SGD achieves better performance, whereas the proposed hybrid method that facilitates GOFAU improves performance by an order of magnitude. The observation that the SGD algorithm performs better on non-convex tasks in a certain setting is not unexpected, as already shown in the literature~\cite{adam_beyond}.

\begin{figure*}
    \centering
    \includegraphics[width=\textwidth]{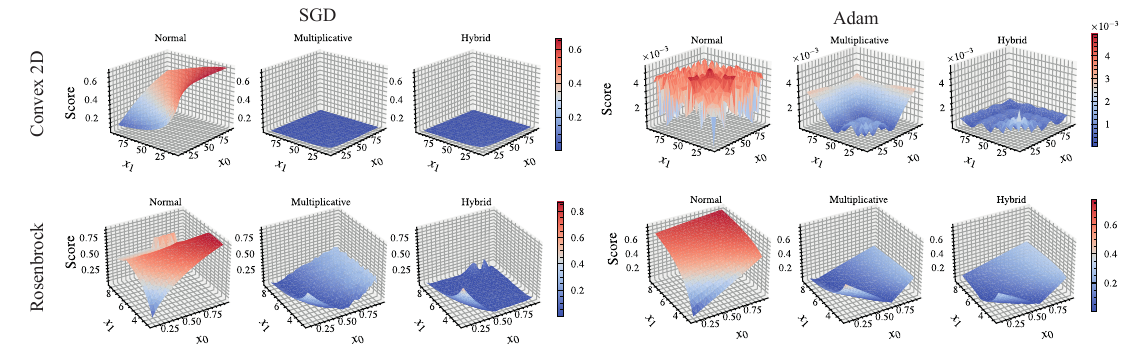}
    \caption{The figure depicts the score of the evaluated update rules applying SGD and Adam optimization methods in two optimization tasks. The z-axis denotes the score in reference to the initial points, given by \(x_{0}\) and \(x_{1}\) as depicted in x and y axis, respectively.}
    \label{fig:surface_plot}
\end{figure*}

Finally, we conducted additional experiments with the SGD and Adam optimizer to evaluate the robustness of each update term. More specifically, using the learning rates obtained from the aforementioned hyperparameter tuning process, we evaluate the performance of optimization methods when different initial points are used in reference to the tuning process, depicting the results in Figure~\ref{fig:surface_plot}. For both tasks, we evaluated the optimizers at 625 different initial points near the initial points used for the tuning process.

As shown in the first row of the figure, the performance of multiplicative updates remains stable in different configurations, without a significant performance degradation occurring when the initial points are distant from the ones used on the tuning process, which is observed when traditional optimizers are used. Although it has been shown in Table~\ref{tab:convex_res} that the proposed method contributes more to the SGD optimizer, we can safely draw the conclusion that it also significantly improves Adam's robustness. As depicted in the second column of Figure~\ref{fig:surface_plot}, the additive Adam has a much rougher surface in contrast to the multiplicative one, especially against the hybrid one. This is even further highlighted in the Rosenbrock task, where the performance variance between additive and multiplicative updates significantly increased, with the hybrid optimizer constantly outperform others irrespective to its initial points, while keeping the performance variance between different configurations relatively low.

\subsection{Image classification}

\begin{table}[]
     \centering
     \caption{Empirically obtained learning rates for the proposed update rules under the GOFAU.}
     \label{tab:learning_rates}
     \begin{tabularx}{\linewidth}{l|XX|XX}
     \toprule
           & \multicolumn{2}{c|}{\textbf{Multiplicative}} & \multicolumn{2}{c}{\textbf{Hybrid} (\(\gamma=0.5\))} \\\cmidrule{2-5}

          Optimizer & \(\eta_{in}\) & \(\eta_{out}\) & \(\eta_{in}\) & \(\eta_{out}\) \\ \midrule 
          SGD & 3.0  & 0.3 & 6.0 & 0.6 \\
          Adagrad & 10 & 0.02 & 8.0 & 0.1 \\
          RMSProp & 0.4 & 0.2 & 0.01 & 0.1 \\
          \bottomrule
     \end{tabularx}
 \end{table}

We conducted additional experiments to evaluate the robustness and convergence of the proposed framework in DNNs applying different size architectures to traditionally used image classification benchmarks. First, we present experimental results in the CIFAR10 dataset applying different training configurations and DNN architectures, such as ResNet9, ResNet18, and VGG16. More specifically, 10 random training configurations are used, applying Xavier initialization\footnote{Xavier initialization randomly draw the weights  from a normal distribution defined as \(\bm{w}\sim\mathcal{N}(0, g\sqrt{\frac{2}{M_i N_i}})\), where \(M_i\) and \(N_i\) are the fan-in and fan-out values of the \(i\)-layer, respectively, and \(g\) denotes the gain} with different gain values and different number of epochs, highlighting in this way the robustness of the proposed multiplicative framework in optimizers SGD, Adagrad, and RMSProp. Gain values are randomly drawn from a gamma distribution, defined as \(g \sim Gamma(k=1, \theta=2.5)\), while the number of training epochs is drawn from a normal distribution, defined as \(t \sim \mathcal{N}(60, 10)\). The default hyperparameter values are used for each optimizer and update rule. Therefore, the learning rates are set to 0.01 for the SGD and Adagrad optimizers when the additive update rule is used and 0.001 for RMSProp, respectively. For the proposed framework, the empirically obtained default values of the learning rates are applied, as presented in Table~\ref{tab:learning_rates}.

In Table~\ref{tab:cifar10_res}, the average evaluation accuracies and their variance are presented in the 10 different training configurations for the evaluated optimization methods. More specifically, in columns 2-4, the validation accuracies at fifth epoch are reported, giving us further insights to the training process, while at columns 5-6 the final validation accuracies are presented. Starting from the SGD optimizer, the significant performance contribution, especially during the first epochs, is highlighted for the proposed GOFAU when applying both multiplicative and hybrid update rules. More precisely, the proposed multiplicative update rule significantly outperforms the traditional additive update, especially in the initial stage of training, with improvements close to \(3\%\) in the cases of ResNet18 and VGG16 at the fifth epoch. Further improvements are observed when the proposed hybrid update rule is applied, highlighting the benefits of multiplicative updates in terms of robustness when there is a high variance in the initial training hypothesis and parameters. Worth mentioning that the proposed hybrid update rule improves ResNet18 performance when the SGD optimizer is used, increasing the evaluation accuracy close to (8\%) in the average case.

Similar results were also obtained when the Adagrad optimizer is used, clearly depicting the acceleration offered when applying multiplicative updates in the results reported for the fifth epoch. More specifically, in the first few epochs, the proposed multiplicative update rule significantly outperforms both evaluated methods, while the proposed hybrid method leads to an overall higher validation accuracy at the end of the training in the average case. Although in the RMSProp optimization method the overall performance contribution of the multiplicative framework is not significantly high at the end of the training, during the first few epochs a significant performance improvement is observed when the hybrid update rule is applied, as reported in the fifth epoch, highlighting its acceleration capabilities.        

\begin{table*}
    \centering
    \caption{Mean and standard deviation of evaluation accuracy applying different optimization methods on 10 randomly drawn training configurations}
    \label{tab:cifar10_res}
    \begin{tabularx}{\textwidth}{X|X|X|X||X|X|X}
         \toprule
         \multirow{2}{*}{\mythead{\normalfont Optimizer}} & \multicolumn{3}{c||}{Epoch 5} & \multicolumn{3}{c}{Final}\\  
         \cmidrule{2-7}
         & \mythead{\normalfont Baseline\\\normalfont(Additive)} & \mythead{\textbf{GOFAU}\\\textbf{(Multiplicative)}} & \mythead{\textbf{GOFAU}\\\textbf{(Hybrid)}}  & \mythead{\normalfont Baseline\\\normalfont(Additive)} & \mythead{\textbf{GOFAU}\\\textbf{(Multiplicative)}} & \mythead{\textbf{GOFAU}\\\textbf{(Hybrid)}} \\
         \midrule
         & \multicolumn{6}{c}{\textbf{ResNet9}}\\ \midrule
         SGD & \(50.33\pm{7.11}\) & \(51.12\pm{8.29}\) & \(\bm{51.38\pm{8.19}}\) & \(76.29\pm{9.68}\) & \(78.10\pm{3.45}\) & \(\bm{79.59\pm{5.14}}\)\\
         Adagrad & \(59.39\pm{5.76}\) & \(\bm{65.68\pm{6.28}}\) & \(62.55\pm{6.59}\) & \(86.68\pm{2.04}\) & \(83.98\pm{2.75}\) & \(\bm{88.56\pm{1.37}}\) \\
         RMSProp & \(61.66\pm{7.47}\) & \(62.64\pm{9.55}\) & \(\bm{64.21\pm{4.85}}\) & \(86.24\pm{4.00}\) & \(85.40\pm{2.18}\) & \(\bm{86.56\pm{2.31}}\)\\
        \midrule
         & \multicolumn{6}{c}{\textbf{ResNet18}}\\ \midrule
         SGD & \(42.80\pm{14.49}\) & \(45.98\pm{13.04}\) & \(\bm{49.60\pm{16.31}}\) & \(73.93\pm{12.95}\) & \(74.81\pm{12.36}\) & \(\bm{81.81\pm{4.16}}\)\\
         Adagrad & \(53.90\pm{5.30}\) & \(\bm{67.86\pm{8.69}}\) & \(59.45\pm{9.99}\) & \(88.09\pm{2.60}\) & \(86.97\pm{1.48}\) & \(\bm{89.02\pm{1.91}}\)\\
         RMSProp & \(61.16\pm{7.02}\) & \(60.15\pm{8.30}\) & \(\bm{63.35\pm{4.82}}\) & \(86.99\pm{3.99}\) & \(86.23\pm{2.18}\) & \(\bm{88.33\pm{2.45}}\)\\
        \bottomrule
         & \multicolumn{6}{c}{\textbf{VGG16}}\\ \midrule
         SGD & \(47.95\pm{15.50}\) & \(51.47\pm{15.28}\) & \(\bm{52.46\pm{16.37}}\) & \(68.91\pm{22.24}\) & \(72.11\pm{22.04}\) & \(\bm{76.05\pm{23.35}}\)\\
         Adagrad & \(28.20\pm{9.37}\) & \(\bm{68.74\pm{3.84}}\) & \(44.71\pm{18.91}\) & \(85.35\pm{5.14}\) & \(85.45\pm{1.47}\) & \(\bm{85.96\pm{2.63}}\)\\
         RMSProp & \(46.49\pm{12.03}\) & \(\bm{57.72\pm{18.24}}\) & \(56.54\pm{6.31}\) & \(85.61\pm{6.28}\) & \(83.00\pm{4.97}\) & \(\bm{85.67\pm{4.56}}\)\\
        \bottomrule
    \end{tabularx}
\end{table*}

\begin{figure}
    \centering
    \includegraphics[width=\linewidth]{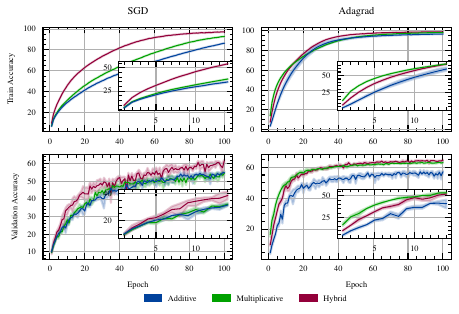}
    \caption{Training and validation accuracy during training applying ResNet18 architecture on CIFAR100 dataset using default configurations for both task and optimizers}
    \label{fig:cifar100}
\end{figure}

Following the results obtained from the CIFAR10 task, we focused on performing experiments using SGD and Adagrad optimizers, since there are cases where greater contributions are observed when the proposed framework is applied. To further understand the acceleration offered by multiplicative updates, we report both the training and validation performance on 5 training runs on large datasets using the default Xavier initialization (\(g=1\)). As depicted in Figure~\ref{fig:cifar100}, hybrid update rule significantly outperforms multiplicative updates, especially during the first 60 epochs. Even in this case where the multiplicative SGD optimizer results in slightly worse performance than the additive SGD, a huge performance boost is observed during training. This is even more clear in the Adagrad optimizer, where applying the multiplicative update leads to faster convergence in the first 20 epochs of training, contrary to the hybrid alternative. Therefore, we can safely draw the conclusion that the benefits of the hybrid update rule can be spotted when exploiting the acceleration capabilities of the multiplicative update rule under GOFAU, while enabling the change sign capabilities of the network by combining with the additive update term.

\begin{figure*}[t!]
    \centering
    \begin{subfigure}[t]{0.49\linewidth}
        \centering
        \includegraphics[width=\linewidth]{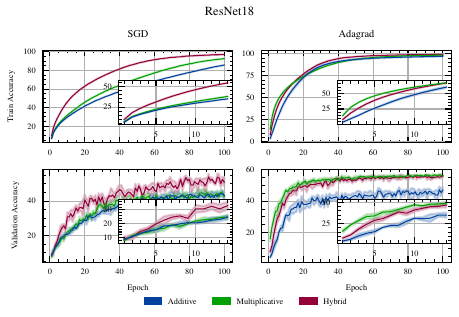}
    \end{subfigure}%
    ~ 
    \begin{subfigure}[t]{0.49\linewidth}
        \centering
        \includegraphics[width=\linewidth]{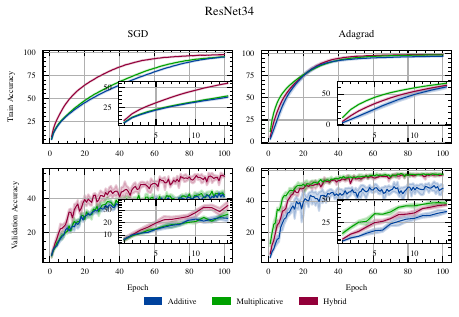}
    \end{subfigure}
    \caption{Training and validation accuracy during training on Tiny ImageNet dataset using default configurations for both task and optimizers}
    \label{fig:large_res}
\end{figure*}

To further highlight the consistency of the observations and the generalizability of the proposed framework, we also performed additional experiments on the Tiny ImageNet dataset applying the ResNet18 and ResNet34 architectures, reported in Figure~\ref{fig:large_res}.The experimental results confirm that the benefits of applying the proposed methods are consistent in both cases where different architectures are used, such as 18- and 34-layer ResNet architectures, and in different datasets, such as the CIFAR100 and Tiny ImageNet datasets, obtaining the corresponding training progress, with Table~\ref{tab:large_res} summarizing the validation accuracy at the end of the training. Furthermore, the generalization ability of the proposed framework is highlighted, starting from a convex task to large DNNs applied on large datasets, with the SGD and Adagrad benefiting more from it. Thus, we conclude that by exploiting the multiplicative updates in an effective way, such in case of the proposed hybrid update rule, one can not only achieve robustness on variant configurations and cases where the theoretical hypothesis difference from the actual one, but also can accelerate the optimization process in cases where a traditional configuration is used.              

\begin{table}[]
    \centering
    \caption{Evaluation performance on CIFAR100 and Tiny ImageNet using default configurations}
    \label{tab:large_res}
    \begin{tabularx}{\linewidth}{l|l|l|l}
        \toprule
         Optimizer & \mythead{\normalfont Baseline\\\normalfont(Additive)} & \mythead{\textbf{GOFAU}\\\textbf{(Multiplicative)}} & \mythead{\textbf{GOFAU}\\\textbf{(Hybrid)}} \\ \midrule
         & \multicolumn{3}{c}{ResNet18 - CIFAR100}\\ \midrule
         SGD & \(55.05\pm{0.68}\) & \(54.51\pm{1.58}\) & \(\bm{61.58\pm{1.10}}\) \\
         Adagrad & \(56.95\pm{1.43}\) & \(62.98\pm{0.44}\) & \(\bm{64.54\pm{0.74}}\) \\ \midrule
         
         & \multicolumn{3}{c}{ResNet18 - Tiny ImageNet}\\ \midrule
         SGD & \(43.30\pm{2.24}\) & \(42.64\pm{1.98}\) & \(\bm{50.97\pm{3.40}}\) \\
         Adagrad & \(46.97\pm{1.84}\) & \(55.75\pm{1.00}\) & \(\bm{55.96\pm{0.54}}\) \\ \midrule
         
         & \multicolumn{3}{c}{ResNet34 - Tiny ImageNet}\\ \midrule
         SGD & \(41.26\pm{1.56}\) & \(42.28\pm{4.23}\) & \(\bm{53.43\pm{1.31}}\) \\
         Adagrad & \(48.28\pm{2.38}\) & \(\bm{57.34\pm{0.73}}\) & \(56.95\pm{0.71}\) \\
         \bottomrule
    \end{tabularx}
\end{table}

\section{Conclusions}
\label{sec:conclusions}

Even though, multiplicative updates hold theoretical claims that potentially can be used to accelerate DL training and lead to robust models, there are merely studied in context of DL optimization. In this work, we propose a novel framework that unlocks training  acceleration and robust model capabilities, exploiting the properties of multiplicative update rules. More specifically, we propose  novel framework enables one to apply alternative update rules during the training, and, in turn, we propose a multiplicative update rule takes into account the magnitude of parameters, while normalize gradients, claiming that leads to faster training and robust models. To overcome limitations that may arise from the multiplicative update rule, we also propose a hybrid method that combines traditional update rule with the proposed multiplicative one. As demonstrated by the conducted evaluation experiments, the proposed framework accelerate training during the initial stage, while it ensures robustness when the actual configurations differs from the initial hypothesis. We validate the proposed framework in a wide range of optimization benchmarks, starting from convex optimization task to traditionally used image classification benchmarks, such as CIFAR and Tiny ImageNet.         

\section*{Acknowledgments}
This project has received funding from the European Union’s Horizon 2020 research and innovation programme under grant agreement No 871391 (PlasmoniAC). This publication reflects the authors' views only. The European Commission is not responsible for any use that may be made of the information it contains.

\bibliographystyle{cas-model2-names}
\bibliography{bibliography}


\end{document}